\pdfoutput=1
\documentclass[11pt]{article}

\usepackage[final]{acl}
\usepackage{amsmath,mathtools}
\usepackage{amsthm}
\newtheorem{proposition}{Proposition}
\usepackage{newtxtext,newtxmath}

\usepackage{tikz}
\usetikzlibrary{positioning}
\tikzset{every edge/.style={draw,auto=right,font={\tiny}}}
\tikzset{input/.style={draw,circle,node contents={}}}
\tikzset{linear/.style={draw,circle,node contents={}}}
\tikzset{relu/.style={draw,circle,inner sep=0mm,minimum size=6mm,node contents={
\begin{tikzpicture}[x=3mm,y=3mm]
\draw (-1,0) -- (0,0) -- (0.707,0.707);
\end{tikzpicture}
}}}
\tikzset{sigmoid/.style={draw,circle,inner sep=0mm,minimum size=6mm,node contents={
\begin{tikzpicture}
\begin{axis}[width=20mm,height=20mm,axis x line=none,axis y line=none]
\addplot[mark=none] {1/(1+exp(-x))};
\end{axis}
\end{tikzpicture}
}}}
\tikzset{att/.style={draw,rectangle,minimum width=1.5cm,minimum height=1cm,node contents={Att},text depth={0.5cm},label={[anchor=east]east:{\tiny Q}},label={[anchor=west]west:{\tiny K}},label={[anchor=south]south:{\tiny V}}}}
\tikzset{every label/.style={font={\small}}}

\usepackage{pgfplots}
\pgfplotsset{compat=1.16}
\usepgfplotslibrary{fillbetween}
\definecolor{red}{HTML}{E5323B}
\definecolor{orange}{HTML}{FFC857}
\definecolor{green}{HTML}{BDD9BF}
\definecolor{blue}{HTML}{A997DF}
\pgfplotsset{
    width=8cm,height=4cm,
    every axis plot/.append style={line width=1pt},
    cycle list={{red},{orange},{green},{blue}}}

\usepackage{graphicx}
\usepackage{booktabs}

\usepackage[capitalise]{cleveref}
\crefformat{section}{#2§#1#3}

\newcommand{\indicator}[1]{\mathbb{I}[#1]}

\newcommand{\avec}{\mathbf{a}}
\newcommand{\qvec}{\mathbf{q}}
\newcommand{\Kmat}{\mathbf{K}}
\newcommand{\Vmat}{\mathbf{V}}
\newcommand{\Wmat}{\mathbf{W}}
\newcommand{\cvec}{\mathbf{c}}
\newcommand{\bvec}{\mathbf{b}}
\newcommand{\xvec}{\mathbf{x}}
\newcommand{\hvec}{\mathbf{h}}

\newcommand{\ffwt}[3][]{#1\Wmat^{\text{F},#2,#3}} % #2=layer, #3=sublayer
\newcommand{\ffbias}[3][]{#1\bvec^{\text{F},#2,#3}} % #2=layer, #3=sublayer
\newcommand{\sawt}[4][]{#1\Wmat^{\text{#4},#2,#3}} % #2=layer, #3=head, #4=Q, K, or V

\newcommand{\att}{\text{Att}}

\newcommand{\onehalf}{\text{\textonehalf}}

\title{Overcoming a Theoretical Limitation of Self-Attention}
\author{David Chiang \and Peter Cholak \\ University of Notre Dame \\ \texttt{\{dchiang,cholak\}@nd.edu}}

\begin{document}
\maketitle
\begin{abstract}
Although transformers are remarkably effective for many tasks, there are some surprisingly easy-looking regular languages that they struggle with. Hahn shows that for languages where acceptance depends on a single input symbol, a transformer's classification decisions become less and less confident (that is, with cross-entropy approaching 1~bit per string) as input strings get longer and longer.
We examine this limitation using two languages: \textsf{PARITY}, the language of bit strings with an odd number of \texttt{1}s, and \textsf{FIRST}, the language of bit strings starting with a \texttt{1}. We demonstrate three ways of overcoming the limitation suggested by Hahn's lemma. First, we settle an open question by constructing a transformer that recognizes \textsf{PARITY} with perfect accuracy, and similarly for \textsf{FIRST}. Second, we use layer normalization to bring the cross-entropy of both models arbitrarily close to zero. Third, when transformers need to focus on a single position, as for \textsf{FIRST}, we find that they can fail to generalize to longer strings; we offer a simple remedy to this problem that also improves length generalization in machine translation.
\end{abstract}

\section{Introduction}
\label{sec:introduction}

Although transformers \citep{vaswani+:2017} are remarkably effective for many tasks, there are some surprisingly easy-looking formal languages that they struggle with. 
\Citet{hahn:2020} tries to explain some of these by showing (his Lemma 5) that changing a single input symbol only changes the output of a transformer encoder by $O(1/n)$, where $n$ is the input string length. 
Thus, for a language where acceptance depends on a single input symbol, a transformer might accept or reject strings with perfect accuracy, but for large $n$, it must do so with low confidence, giving accepted strings a probability just above $\onehalf$ and rejected strings a probability just below $\onehalf$. More precisely, as $n$ increases, the cross-entropy approaches its worst possible value of 1~bit per string.

Here, we examine this limitation using two simple regular languages:
\begin{align*}
\textsf{PARITY} &= \{w \in \Sigma^\ast \mid \text{$w$ has an odd number of \texttt{1}s}\} \\
\textsf{FIRST} &= \{ w \in \Sigma^\ast \mid w_1 = \texttt{1}\}
\end{align*}
where (here and throughout the paper) $\Sigma = \{\texttt{0}, \texttt{1}\}$.
Hahn's lemma applies to \textsf{PARITY} because the network must attend to all the symbols of the string, and a change in any one of them changes the correct answer.
We have chosen \textsf{FIRST} as one of the simplest examples of a language that the lemma applies to. It only requires attention on the first symbol, but the lemma still applies because a change in this symbol changes the correct answer.

Although the lemma might be interpreted as limiting the ability of transformers to recognize these languages, we show three ways that this limitation can be overcome.

First, we show by explicit constructions that transformers do in fact exist that can recognize both languages with perfect accuracy for arbitrary lengths. We have implemented these constructions and verified them experimentally~(\cref{sec:exact}).

As predicted by Hahn's lemma, our constructed transformers have cross-entropy that approaches 1~bit (that is, just barely better than random guessing) as input length increases. But we show that by adding layer normalization, the cross-entropy can be made arbitrarily close to zero, independent of string length (\cref{sec:layernorm}).

In practice, we find, like \citet{bhattamishra+:2020}, that transformers cannot learn \textsf{PARITY}. Perhaps more surprisingly, when learning \textsf{FIRST}, transformers can have difficulty generalizing from shorter strings to longer strings. Although this is not a logical consequence of Hahn's lemma, it is a consequence of the behavior that Hahn's lemma predicts. Fortunately, this problem can be fixed with a simple modification, multiplying attention logits by $\log n$. This modification also improves length generalization in machine translation~(\cref{sec:learnability}).

\section{Background}

\subsection{Notation}

If $\phi$ is a true-or-false statement, we write
\begin{equation*}
\indicator{\phi} = \begin{cases} 1 & \text{if $\phi$ is true} \\
0 & \text{otherwise.}
\end{cases}
\end{equation*}

For any $m, n > 0$, we write $\mathbf{0}^{m \times n}$ for the $m\times n$ zero matrix and $\mathbf{I}^{n \times n}$ for the $n \times n$ identity matrix.

\subsection{Transformers}

Following \citet{hahn:2020}, we consider transformer encoders with a sigmoid output layer on a single position. Differently from \citet{hahn:2020}, but in line with common practice \citep{devlin+:2019}, we prepend a token \texttt{CLS} (for ``classification'') and use the encoder output at this token's position for classifying the string.

We use the original definition of transformers \citep{vaswani+:2017}, except for positional encodings.

\subsubsection{Input layer}

The input to the network is a string $w \in \Sigma^*$. 
Let $n = |w|+1$, let $w_0 = \texttt{CLS}$, and let $w_i$ be the $i$-th symbol of~$w$. 

The input layer has a word embedding and positional encodings,
\begin{align*}
\text{WE} \colon \Sigma &\rightarrow \mathbb{R}^d \\
\text{PE} \colon \mathbb{N} &\rightarrow \mathbb{R}^d
\end{align*}
which are used to compute input vectors for $i=0, \ldots n$:
\begin{align*}
\avec^{0,i} &= \text{WE}(w_i) + \text{PE}(i).
\end{align*}
The word embeddings are typically learned, while the positional encodings vary somewhat. Originally \citep{vaswani+:2017}, they were fixed and defined in terms of sine and cosine waves, but they can also be learned \citep{gehring+:2017}, in which case they are defined only up to some maximum position. Here, we allow ourselves to define $\text{PE}$ as an arbitrary function on all positions. It would seem that to remain in the spirit of the original paper, $\text{PE}$ should be easy to compute, independent of $w$, and parallelizable over positions.

\subsubsection{Encoder layers}

The body of the encoder is a stack of $L$ layers, each of which has a self-attention sublayer followed by a position-wise feedforward sublayer.
For $\ell=1, \ldots, L$, layer $\ell$ is defined as follows, where $h=1, \ldots, H$, and $i=0, \ldots, n$:
\begin{align*}
\qvec^{\ell,h,i} &= \sawt{\ell}hQ \avec^{\ell-1,i} \\
\Kmat^{\ell,h} &= \begin{bmatrix} \sawt{\ell}hK \avec^{\ell-1,0} & \cdots & \sawt{\ell}hK \avec^{\ell-1,n} \end{bmatrix}^\top \\
\Vmat^{\ell,h} &= \begin{bmatrix} \sawt{\ell}hV \avec^{\ell-1,0} & \cdots & \sawt{\ell}hV \avec^{\ell-1,n} \end{bmatrix}^\top \\
\cvec^{\ell,i} &= \text{LN}\left(\sum_{h=1}^H \att(\qvec^{\ell,h,i}, \Kmat^{\ell,h}, \Vmat^{\ell,h}) + \avec_{\ell-1,i} \right) \\
\hvec^{\ell, i} &= \max\left(0, \ffwt\ell1 \cvec^{\ell,i} + \ffbias\ell1 \right) \\
\avec^{\ell, i} &= \text{LN}\left(\ffwt\ell2 \hvec^{\ell,i} + \ffbias\ell2 + \cvec^{\ell,i}\right)
\end{align*}
where boldface lowercase letters stand for vectors in $\mathbb{R}^d$ and boldface uppercase letters stand for matrices in $\mathbb{R}^{d \times d}$. The learned parameters of the model are the $\mathbf{W}$'s and $\mathbf{b}$'s.
The function $\att$ is scaled dot-product attention, defined as
\begin{gather*}
\att \colon \mathbb{R}^{d} \times \mathbb{R}^{(n+1)\times d} \times \mathbb{R}^{(n+1) \times d} \rightarrow \mathbb{R}^d \\
\att(\mathbf{q}, \mathbf{K}, \mathbf{V}) = \mathbf{V}^\top \operatorname{softmax} \frac{\mathbf{K}\mathbf{q}}{\sqrt{d}}
\end{gather*}
where the result of the $\operatorname{softmax}$, sometimes written as $\alpha$, is a vector of \emph{attention weights}.
The function $\text{LN}$ is layer normalization, whose definition we defer to \cref{sec:layernorm}.

\subsubsection{Output layer}

Finally, the network linearly projects the encoding of $\texttt{CLS}$ to a scalar and applies a sigmoid function:
\begin{align*}
y &= \sigma(\Wmat^{L+1} \avec^{L,0} + \bvec^{L+1})
\end{align*}
where $\Wmat^{L+1} \in \mathbb{R}^{1 \times d}$ and $\bvec^{L+1} \in \mathbb{R}^{1 \times 1}$. The network accepts $w$ iff the output probability is greater than $\tfrac12$.

\section{Exact Solutions}
\label{sec:exact}

The first way to overcome the limitation suggested by Hahn's lemma is to show by explicit construction that our two languages can in fact be recognized with perfect accuracy by transformers.

\subsection{FFNN for \textsf{PARITY}}

\citet{rumelhart+:1986} showed that for any $n$, there is a feedforward neural network (FFNN) that computes \textsf{PARITY} for strings of length exactly $n$. They also showed that a randomly initialized FFNN can learn to do this automatically.

Since our construction is partially based on theirs, it may be helpful to review their construction in detail. Let $w$ be the input string, $|w| = n$, and $k$ be the number of $\texttt{1}$s in $w$. The input is a vector $\xvec$ such that $\xvec_i = \indicator{w_i = \texttt{1}}$. The first layer computes $k$ and compares it against $1, 2, \ldots, n$:
\begin{align*}
\Wmat^1 &= \begin{bmatrix}
1 & 1 & \cdots & 1\\
1 & 1 & \cdots & 1\\
\vdots & \vdots & \ddots & \vdots \\
1 & 1 & \cdots & 1
\end{bmatrix}
&
\bvec^1 &= \begin{bmatrix}
-0.5 \\
-1.5 \\
\vdots \\
-n+0.5 \\
\end{bmatrix}
\end{align*}
so that
\begin{align*}
\hvec^1 &= H(\Wmat^1 \xvec + \bvec^1) = \begin{bmatrix}
\indicator{k \geq 1} \\
\indicator{k \geq 2} \\
\vdots \\
\indicator{k \geq n}
\end{bmatrix}
\end{align*}
where $H$ is the step function ($H(x) = \indicator{x > 0}$), applied elementwise.

The second layer adds up the odd elements and subtracts the even elements:
\begin{align*}
\mathbf{W}^2 &= \begin{bmatrix} 1 & -1 & \cdots & (-1)^{n+1} \end{bmatrix} & \bvec^2 &= -0.5  \\
y &= H(\mathbf{W}^2 \hvec^1 + \bvec^2)
\end{align*}
which is $1$ if $k$ is odd and $0$ is $k$ is even.

\subsection{Transformer for \textsf{PARITY}}
\label{sec:parity_exact}

\begin{proposition}
There is a transformer encoder with sigmoid output layer that recognizes (in the above sense) the language $\textsf{PARITY}$ for strings of arbitrary length.
\end{proposition}

Initially, we will construct a transformer encoder without layer normalization (that is, $\text{LN}(\xvec) = \xvec$); then we will show how to add layer normalization~(\cref{sec:layernorm}). 
Let $k$ be the number of occurrences of $\texttt{1}$ in~$w$.
All vectors computed by the network have $d=9$ dimensions; if we show fewer dimensions, assume the remaining dimensions to be zero.

\iffalse
\begin{figure}
\centering
\input{parity_figure}
\caption{Diagram of transformer recognizing \textsf{PARITY}. Residual connections and layer normalization are not shown.}
\label{fig:parity}
\end{figure}
\fi

The word and position embeddings are:
\begin{align*}
  \text{WE}(\texttt{0}) &= \begin{bmatrix} 1 \\ 0 \\ 0 \\ 0 \\ 0 \end{bmatrix} &
  \text{WE}(\texttt{1}) &= \begin{bmatrix} 0 \\ 1 \\ 0 \\ 0 \\ 0 \end{bmatrix} \\ 
  \text{WE}(\texttt{CLS}) &= \begin{bmatrix} 0 \\ 0 \\ 1 \\ 0 \\ 0 \end{bmatrix} &
  \text{PE}(i) &= \begin{bmatrix} 0 \\ 0 \\ 0 \\ \frac{i}{n} \\ \cos i\pi \end{bmatrix}.
\end{align*}
Since we are numbering positions starting from 0, dimension 4 ranges from 0 to $\frac{n-1}{n}$, and dimension~5 is $+1$ for even positions and $-1$ for odd positions.

We argue that dimension 5, being a cosine wave, is a fairly standard choice, although its period (2) is shorter than the shortest period in standard sinusoidal encodings ($2\pi$).
Dimension 4 is admittedly not standard; however, we argue that it is a reasonable encoding, and extremely easy to compute.

Thus, the encoding of word $w_i$ is:
\begin{equation*}
  \avec^{0,i}
  %= \text{WE}(w_i)+\text{PE}(i) 
  = \begin{bmatrix} \indicator{w_i=\texttt{0}} \\ \indicator{w_i=\texttt{1}} \\ \indicator{w_i=\texttt{CLS}} \\ \frac{i}{n} \\ \cos i\pi \end{bmatrix}.
\end{equation*}

The network has $L=2$ layers and $H=2$ heads.
The first self-attention layer has one head which finds $k$, the number of $\texttt{1}$s. 
More precisely, because attention always averages, it must compute the ``average'' number of $\texttt{1}$s, that is, $\frac{k}{n}$, and stores it in dimension 6. 
It also stores $\frac{1}{n}$ in dimension 7, which we will need later.
\begin{align*}
  \sawt11Q &= \mathbf{0} \\
  \sawt11K &= \mathbf{0} \\
  \sawt11V &=
  \begin{bmatrix}
    \mathbf{0}^{5\times 5} \\
    \begin{matrix}
    0 & 1 & 0 & 0 & 0 \\
    0 & 0 & 1 & 0 & 0
    \end{matrix}
  \end{bmatrix}
\end{align*}
The second head doesn't do anything ($\sawt12V = \mathbf{0}$; the queries and keys can be anything).
After the residual connection, we have:
\begin{align*}
  \cvec^{1,i} &= \begin{bmatrix} \indicator{w_i=\texttt{0}} \\ \indicator{w_i=\texttt{1}} \\ \indicator{w_i=\texttt{CLS}} \\ \frac{i}{n} \\ \cos i\pi \\ \frac{k}{n} \\ \frac{1}{n} \end{bmatrix}.
\end{align*}

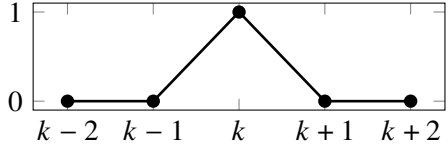
\begin{figure}
\centering
\begin{tikzpicture}
\begin{axis}[width=7cm,height=3cm,ytick={0,1},xtick={-2,-1,0,1,2},xticklabels={$k-2$,$k-1$,$k$,$k+1$,$k+2$,$k+3$}]
\addplot[mark=*] coordinates {(-2,0) (-1,0) (0,1) (1,0) (2,0) };
\end{axis}
\end{tikzpicture}
\caption{Piecewise linear function equivalent on the integers to $\indicator{i=k}$.}
\label{fig:piecewise}
\end{figure}

In the construction of \citet{rumelhart+:1986}, the next step is to compute $\indicator{i\leq k}$ for each $i$, using step activation functions. 
There are two differences in our construction. First, we have ReLU activation functions, not step activation functions. 
Second, because attention must sum to one, if $n$ is odd then the even and odd positions will get different attention weights, so the trick of subtracting even positions from odd positions will not work. Instead, we want to compute $\indicator{i = k}$ (\cref{fig:piecewise}).

The first FFNN has two layers. The first is:
\begin{align*}
  \ffwt11 &= \begin{bmatrix}
    0 & 0 & 0 & -1 & 0 & 1 & -1 \\
    0 & 0 & 0 & -1 & 0 & 1 & 0 \\
    0 & 0 & 0 & -1 & 0 & 1 & 1
  \end{bmatrix} \\
  \ffbias11 &= \begin{bmatrix}
    0 \\
    0 \\
    0
  \end{bmatrix}.
\end{align*}
This gives:
\begin{align*}
  \hvec^{1,i} &= \frac1n
  \begin{bmatrix} 
    \max(0,k-i-1) \\ 
    \max(0,k-i) \\ 
    \max(0,k-i+1) 
  \end{bmatrix}.
\end{align*}
The second layer linearly combines these three values to get $\indicator{i = k}$ as desired.
\begin{align*}
  \ffwt12 &= \begin{bmatrix}
    \mathbf{0}^{7\times 3} \\
    \begin{matrix} 1 & -2 & 1 \end{matrix}
  \end{bmatrix} &
  \ffbias12 &= \mathbf{0}.
\end{align*}
After the residual connection, we have:
\begin{align*}
  \avec^{1,i} &= \begin{bmatrix} \indicator{w_i=\texttt{0}} \\ \indicator{w_i=\texttt{1}} \\ \indicator{w_i=\texttt{CLS}} \\ \frac{i}{n} \\ \cos i\pi \\ \frac{k}{n} \\ \frac{1}{n} \\ \frac{\indicator{i = k}}{n} \end{bmatrix}.
\end{align*}

The second self-attention layer tests whether position $k$ is even or odd. It does this using two heads, one which attends more strongly to the odd positions, and one which attends more strongly to the even positions; both average dimension 8:
\begin{align*}
  \sawt21Q &= \begin{bmatrix} 0 & 0 & c\sqrt{d} & 0 & 0 & 0 & 0 & 0  \end{bmatrix} \\
  \sawt21K &= \begin{bmatrix} 0 & 0 & 0 & 0 & -1 & 0 & 0 & 0 \end{bmatrix} \\
  \sawt21V &=
  \begin{bmatrix}
  \mathbf{0}^{8 \times 8} \\
    \begin{matrix}
    0 & 0 & 0 & 0 & 0 & 0 & 0 & 1
    \end{matrix}
  \end{bmatrix} \\
  \sawt22Q &= \begin{bmatrix} 0 & 0 & c\sqrt{d} & 0 & 0 & 0 & 0 & 0 \end{bmatrix} \\
  \sawt22K &= \begin{bmatrix} 0 & 0 & 0 & 0 & 1 & 0 & 0 & 0 \end{bmatrix} \\
  \sawt22V &= 
  \begin{bmatrix}
  \mathbf{0}^{8 \times 8} \\
    \begin{matrix}
    0 & 0 & 0 & 0 & 0 & 0 & 0 & -1
    \end{matrix}
  \end{bmatrix}
\end{align*}
where $c > 0$ can be any constant. 
The second FFNN doesn't do anything ($\ffwt21 = \ffbias21 = \ffwt22 = \ffbias22 = \mathbf{0}$). The vector at \texttt{CLS} (position $0$) is then
\begin{align*}
  \avec^{2,0} &= 
\begin{bmatrix} 0 \\ 0 \\ 1 \\ 0 \\ 1 \\ \frac{k}{n} \\ \frac{1}{n} \\ \frac{\indicator{k = 0}}{n} \\
  s
\end{bmatrix}
\end{align*}
where $s$ has a somewhat complicated value.
If $n$ is even, it turns out to be
\begin{equation*}
s = (-1)^{k+1} \frac{2 \tanh c}{n^2}
\end{equation*}
which is positive if $k$ is odd and negative if $k$ is even. As predicted by Hahn, it is in $O(1/n)$. If $n$ is odd, the expression for $s$ is more complicated (see~\cref{app:parity_details}), but it is still positive iff $k$ is odd, and it is still in $O(1/n)$.

Finally, the output layer is a sigmoid layer that just looks at dimension 9:
\begin{align*}
  \Wmat^3 &= \begin{bmatrix}
    0 & 0 & 0 & 0 & 0 & 0 & 0 & 0 & 1
    \end{bmatrix} & \bvec^3 = \mathbf{0} \\
    y &= \frac1{1+\exp(-s)}.
\end{align*}
So the output is greater than $\tfrac12$ iff $k$ is odd.

\subsection{Transformer for \textsf{FIRST}}
\label{sec:first_exact}

Next, we construct a transformer for \textsf{FIRST}.
In line with the common practice of learning per-position word embeddings \citep{gehring+:2017}, we use position embeddings that test whether a word is at position 1:
\begin{align*}
\avec^{0,i} &= \begin{bmatrix} \indicator{w_i = \texttt{0}} \\ 
\indicator{w_i = \texttt{1}} \\ 
\indicator{w_i = \texttt{CLS}} \\ 
\indicator{i=1} 
\end{bmatrix}.
\end{align*}
The first self-attention layer does nothing ($\sawt11V = \mathbf{0}$), so after the residual connection, $\cvec^{1,i} = \avec^{0,i}$.

The first FFNN computes a new component (5) that tests whether $i=1$ and $w_1 = \texttt{1}$:
\begin{align*}
\ffwt11 &=
\begin{bmatrix}
-1 & 0 & -1 & 1
\end{bmatrix} & \ffbias11 &= 0 \\
\ffwt12 &= 
\begin{bmatrix}
0 \\ 0 \\ 0 \\ 0 \\ 1
\end{bmatrix}
& \ffbias12 &= \mathbf{0} \\
\avec^{1,i} &= \begin{bmatrix} \indicator{w_i = \texttt{0}} \\ 
\indicator{w_i = \texttt{1}} \\ 
\indicator{w_i = \texttt{CLS}} \\ 
\indicator{i=1} \\
\indicator{w_i = \texttt{1} \land i=1}
\end{bmatrix}.
\end{align*}
(We have chosen $\ffwt11$ in a slightly unusual way to avoid using the bias term $\ffbias11$, in anticipation of \cref{sec:layernorm} when we will add layer normalization.)

The second self-attention layer has a single head, which makes \texttt{CLS} focus on position 1.
\begin{align*}
\sawt21Q &= 
\begin{bmatrix}
0 & 0 & c\sqrt{d} & 0 & 0
\end{bmatrix}
\\
\sawt21K &=
\begin{bmatrix}
0 & 0 & 0 & 1 & 0
\end{bmatrix}
\\
\sawt21V &=
\begin{bmatrix}
\mathbf{0}^{5\times 5} \\
\begin{matrix}
0 & 0 & 0 & -\tfrac12 & 1
\end{matrix}
\end{bmatrix}
\end{align*}
where $c > 0$ is a constant.
The second FFNN doesn't do anything ($\ffwt21 = \ffbias21 = \ffwt22 = \ffbias22 = \mathbf{0}$). So at \texttt{CLS} (position $0$),
\begin{align}
\avec^{2,0} &= 
\begin{bmatrix}
0 \\
0 \\
1 \\
0 \\
0 \\
s
\end{bmatrix} \notag \\
s &= \frac{\exp c}{\exp c+n-1}\left(\indicator{w_1 = \texttt{1}} - \tfrac12\right). \label{eq:first_exact_logit}
\end{align}
The final output layer just selects component 6:
\begin{align*}
\Wmat^3 &= \begin{bmatrix}
0 & 0 & 0 & 0 & 0 & 1
\end{bmatrix}
& \bvec^3 &= 0.
\end{align*}
So the output probability, $y = \sigma(s)$, is greater than $\tfrac12$ iff $w_1 = \texttt{1}$. However, it will get closer to $\tfrac12$ as $n$ increases.

\subsection{Experiments}

We implemented both of the above constructions using modified versions of PyTorch's built-in implementation of transformers \citep{pytorch}.\footnote{The code for this and other experiments in this paper are available at \url{https://github.com/ndnlp/parity}.} These constructions achieve perfect accuracy for strings with lengths sampled from $[1, 1000]$.

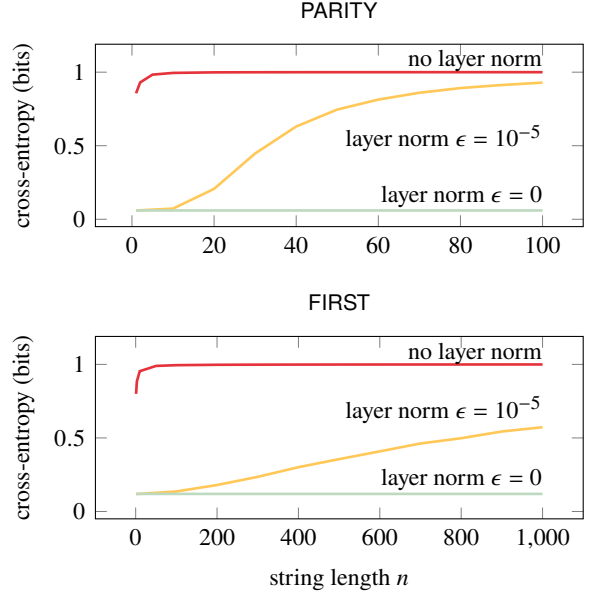
\begin{figure}
\pgfplotsset{every axis/.append style={font={\small}}}
\centering
\begin{tikzpicture}
  \begin{axis}[title={\textsf{PARITY}},%xlabel={string length $n$},
  ylabel={cross-entropy (bits)},ymin=-0.05,ymax=1.19]
    % no layernorm
    \addplot coordinates {
      (1,  0.8565826382974341)
      (2,  0.9301572712306551)
      %(3,  0.962660849419011 )
      %(4,  0.9764992837957805)
      (5,  0.9832823431793906)
      %(6,  0.9877607937656028)
      %(7,  0.9905722577928379)
      %(8,  0.9923973462829265)
      %(9,  0.9939590261449538)
      (10, 0.9949348995615253)
      (20, 0.9986299786321032)
      (30, 0.9993688126605568)
      (40, 0.9996391229523974)
      (50, 0.9997660392694282)
      (60, 0.9998375737332279)
      (70, 0.9998802512280396)
      (80, 0.9999078234866258)
      (90, 0.9999270571663852)
      (100,0.9999406549746755)
    };
    \node[anchor=south east,inner sep=0pt] at (axis cs:100,1) {no layer norm};

    % eps=1e-5
    \addplot coordinates {
      (1,  0.0598268071861018)
      %(2,  0.0598982892489378)
      %(5,  0.0609157173285891)
      (10, 0.0723226858712001)
      %(15, 0.1160454334055839)
      (20, 0.2067803639976356)
      (30, 0.4473115602690626)
      (40, 0.6307018427213459)
      (50, 0.7458586989541531)
      (60, 0.8136844261111903)
      (70, 0.8600002551776352)
      (80, 0.8919408928738809)
      (90, 0.9125824415311775)
      (100,0.9289497647005954)
    };
    \node[anchor=north east,inner sep=0pt] at (axis cs:100,0.65) {layer norm $\epsilon=10^{-5}$};

    % eps=0
    \addplot coordinates {
      (1,  0.0598109869320056)
      (10, 0.0598109869320056)
      (20, 0.0598110235858080)
      (30, 0.0598110436325358)
      (40, 0.0598112205059433)
      (50, 0.0598110779753214)
      (60, 0.0598110922713792)
      (70, 0.0598111822398035)
      (80, 0.0598113170849507)
      (90, 0.0598114690746184)
      (100,0.0598109033591862)
    };
    \node[anchor=south east,inner sep=0pt] at (axis cs:100,.08) {layer norm $\epsilon=0$};
  \end{axis}
\end{tikzpicture}
\\[0.25cm]
\begin{tikzpicture}
  \begin{axis}[title={\textsf{FIRST}},xlabel={string length $n$},ylabel={cross-entropy (bits)},ymin=-0.05,ymax=1.19]
    \addplot coordinates {
      % for I in $(seq 10 10 100); do python3 first_exact.py --steps 1000 --length $I --big 1; done
      (1,  0.7995605865253138)
      %(2,  0.8534708702040478)
      (3,  0.884574362569214 )
      %(4,  0.9047996083020057)
      %(5,  0.9189990978965171)
      %(6,  0.9295151490661745)
      %(7,  0.9376157039044636)
      %(8,  0.9440469091421528)
      %(9,  0.9492763856069383)
      (10, 0.953612326209166 )
      %(20, 0.9749920914771998)
      %(30, 0.9828822255421583)
      %(40, 0.9869877953835553)
      (50, 0.9895050194529087)
      %(60, 0.9912060998525928)
      %(70, 0.9924327661098717)
      %(80, 0.9933589786760935)
      %(90, 0.9940832836102022)
      (100,0.9946650149267456)
      (200, 0.997310194088341)
      (300, 0.9982017521504138)
      (400, 0.9986495089819352)
      (500, 0.9989186618305355)
      (600, 0.999098383700687)
      (700, 0.999226940732207)
      (800, 0.9993232510166902)
      (900, 0.9993983214437918)
      (1000,0.9994584293802684)
    };
    \node[anchor=south east,inner sep=0pt] at (axis cs:1000,1) {no layer norm};

    \addplot coordinates {
      % for I in $(seq 10 10 100); do python3 first_exact_layernorm.py --steps 1000 --length $I --big 1; done
      (1,   0.1194991385555496)
      (10,  0.11974176126368756)
      %(20,  0.12031935001262185)
      %(30,  0.12125278395837058)
      %(40,  0.1225127060923881 )
      %(50,  0.12392733553170783)
      %(60,  0.12589008137555094)
      %(70,  0.12814158014604235)
      %(80,  0.1306710256035358 )
      %(90,  0.13304103999424713)
      (100, 0.13620344581018906)
      (200, 0.179861393818121  )
      (300, 0.23549239600425068)
      (400, 0.30119035434225894)
      (500, 0.3558859009588258 )
      (600, 0.4086388846373034 )
      (700, 0.4623186103950869 )
      (800, 0.49803669412495755)
      (900, 0.5437462754900045 )
      (1000,0.5727302169476414 )
    };
    \node[anchor=south east,inner sep=0pt] at (axis cs:1000,0.6) {layer norm $\epsilon=10^{-5}$};

    \addplot coordinates {
      % for I in $(seq 10 10 100); do python3 first_exact_layernorm.py --steps 1000 --length $I --big 1 --eps 0; done
      (1,   0.11947247794474875)
      (100, 0.11947262842956825)
      (200, 0.11947)
      (300, 0.11947)
      (400, 0.11947)
      (500, 0.11947)
      (600, 0.11947)
      (700, 0.11947)
      (800, 0.11947)
      (900, 0.11947)
      (1000,0.11947255348812814)
    };
    \node[anchor=south east,inner sep=0pt] at (axis cs:1000,.14) {layer norm $\epsilon=0$};

  \end{axis}
\end{tikzpicture}
\vspace*{-1ex}
\caption{Cross-entropy of exact solutions for \textsf{PARITY} and \textsf{FIRST} computed over 1000 random strings of length $n$. Without layer norm, the cross-entropy quickly approaches its upper bound of one. With layer norm and $\epsilon > 0$, the cross-entropy is better but still grows with $n$. With $\epsilon = 0$, cross-entropy is independent of $n$.}
\label{fig:exact_graphs}
\end{figure}

However, in \cref{fig:exact_graphs}, the red curves (``no layer norm'') show that, as strings grow longer, the cross-entropy approaches its worst possible value of 1~bit per string. We discuss this problem next.

\section{Layer Normalization}
\label{sec:layernorm}

The second way to mitigate or eliminate the limitation of Hahn's lemma is layer normalization \citep{ba+:2016}, which is defined, for any vector~$\xvec$, as
\begin{align*}
\text{LN}(\mathbf{x}; \gamma, \beta) &= \frac{\mathbf{x} - \text{mean}(\mathbf{x})}{\sqrt{\text{var}(\mathbf{x}) + \epsilon}} \circ \gamma + \beta
\end{align*}
where the functions $\text{mean}$ and $\text{var}$ compute the mean and variance, respectively, of the elements of $\mathbf{x}$, and $\circ$ is the elementwise (Hadamard) product. We fix $\beta = 0$ and $\gamma = 1$, so that the result has approximately zero mean and unit variance. The constant $\epsilon$ was not present in the original definition \citep{ba+:2016} but is added in all implementations that we are aware of, for numerical stability.

The original transformer definition performs layer normalization immediately after every residual connection.\footnote{It is also common to place layer normalization before residual connections \citep{wang+:2019,nguyen+salazar:iwslt2019}, but we follow the original transformer definition here.} 
In this section, we modify our two constructions above to use layer normalization. This modification has two steps.

\subsection{Removing centering}
\label{sec:recentering}

The first is to nullify the centering effect of layer normalization by making the network compute each value $x$ as well as its negation $-x$. The new word encodings are defined in terms of those in the original construction:
\begin{align*}
\bar\avec^{0,i} &= \begin{bmatrix}
\avec^{0,i} \\
-\avec^{0,i}
\end{bmatrix}. \\
\intertext{Likewise for the self-attention parameters:}
\sawt[\bar]{\ell}hQ &= \begin{bmatrix}
\sawt{\ell}hQ & \mathbf{0}
\end{bmatrix} \\
\sawt[\bar]{\ell}hK &= \begin{bmatrix}
\sawt{\ell}hK & \mathbf{0}
\end{bmatrix} \\
\sawt[\bar]{\ell}hV &= \begin{bmatrix}
\sawt{\ell}hV & \mathbf{0} \\
-\sawt{\ell}hV & \mathbf{0}
\end{bmatrix}. \\
\intertext{Likewise for the position-wise FFNN parameters:}
\ffwt[\bar]\ell1 &= \begin{bmatrix}
\ffwt\ell1 & \mathbf{0}
\end{bmatrix} & \ffbias[\bar]\ell1 &= \ffbias\ell1 \\
\ffwt[\bar]\ell2 &= \begin{bmatrix}
\ffwt\ell2 \\ -\ffwt\ell2
\end{bmatrix} & \ffbias[\bar]\ell2 &= \begin{bmatrix}
\ffbias\ell2 \\
-\ffbias\ell2
\end{bmatrix}.
\end{align*}
Then each layer of activations is
\begin{align*}
\bar\cvec^{\ell,i} &= \text{LN}\left(\begin{bmatrix}
\cvec^{\ell,i} \\ -\cvec^{\ell,i}
\end{bmatrix}\right)
&
\bar\avec^{\ell,i} &= \text{LN}\left(\begin{bmatrix}
\avec^{\ell,i} \\ -\avec^{\ell,i}
\end{bmatrix}\right).
\end{align*}
The argument to $\text{LN}$ always has zero mean, so that layer normalization does not add or subtract anything. It does scale the activations, but in the case of the two transformers constructed above, any activation layer can be scaled by any positive number without changing the final decisions (see \cref{app:scale}).

\subsection{Reducing cross-entropy}

Furthermore, in any transformer, we can use layer normalization to shrink the cross-entropy as small as we like, contrary to Hahn's Lemma~5. In Hahn's formulation, position-wise functions like layer normalization can be subsumed into his $f^\text{act}$, but the lemma assumes that $f^\text{act}$ is Lipschitz-continuous, and layer normalization with $\epsilon=0$ is not.

\begin{proposition} \label{thm:layernorm}
For any transformer $T$ with layer normalization ($\epsilon=0$) that recognizes a language $\mathcal{L}$, and for any $\eta > 0$, there is a transformer with layer normalization that recognizes $\mathcal{L}$ with cross-entropy at most $\eta$.
\end{proposition}

\begin{proof}
Let $d$ be the number of dimensions in the original vectors of activations, and
let $L$ be the number of layers. Then we add a new layer whose self-attention doesn't do anything ($\sawt{L+1}hV = \mathbf{0}$) and whose FFNN is defined in terms of the original output layer:
\begin{align*}
\ffwt[\bar]{L+1}1 &= \begin{bmatrix}
\mathbf{I}^d \\
-\mathbf{I}^d
\end{bmatrix}
\qquad \ffbias[\bar]{L+1}1 = \begin{bmatrix}
\mathbf{0}^d \\
\mathbf{0}^d
\end{bmatrix}
\\
\ffwt[\bar]{L+1}2 &= \begin{bmatrix}
-\mathbf{I}^d & \mathbf{I}^d
\end{bmatrix} + \begin{bmatrix}
\Wmat^{L+1} & -\Wmat^{L+1} \\
-\Wmat^{L+1} & \Wmat^{L+1} \\
\mathbf{0}^{(d-2) \times d} & \mathbf{0}^{(d-2) \times d}
\end{bmatrix}
\\ 
\ffbias[\bar]{L+1}2 &= \begin{bmatrix}
\bvec^{L+1} \\
-\bvec^{L+1} \\
\mathbf{0}^{d-2}
\end{bmatrix}.
\end{align*}
This causes the residual connection to zero out all dimensions except two, so that if $s$ was the original output logit, the output of this new layer (before layer normalization) is
\begin{equation*}
\bar\avec^{L+1,i} = \text{LN}\left(\begin{bmatrix}
s \\
-s \\
\mathbf{0}^{d-2}
\end{bmatrix}\right).
\end{equation*}
Now, if $\epsilon=0$, layer normalization scales this vector to have unit variance exactly, so it becomes
\begin{equation*}
\bar\avec^{L+1,i} = \begin{bmatrix}
\pm\sqrt{d/2} \\
\mp\sqrt{d/2} \\
\mathbf{0}^{d-2}
\end{bmatrix}.
\end{equation*}

The new output layer simply selects the first dimension, scaling it by $c$:
\begin{align*}
\bar\Wmat^{L+2} &= \begin{bmatrix} c & 0 & \mathbf{0}^{d-2} \end{bmatrix} & \bar\bvec^{L+2} = 0.
\end{align*}
Finally, set $c = -\frac{1}{\sqrt{d/2}} \log (\exp \eta - 1)$. If the input string is in $\mathcal{L}$, then the cross-entropy is $\log \sigma(c\sqrt{d/2}) = \eta$. Similarly, if the input string is not in $\mathcal{L}$, then the cross-entropy is $\log (1-\sigma(-c\sqrt{d/2})) = \eta$.
\end{proof}

However, in practice, $\epsilon$ is always set to a nonzero value, which makes layer normalization Lipschitz-continuous, so Hahn's Lemma 5 still applies.

\subsection{Experiments}

We tested our exact solutions, modified as described above to use layer normalization. \Cref{fig:exact_graphs} shows that layer normalization with $\epsilon>0$ improves the cross-entropy, but it still grows with $n$ and approaches~$1$. With $\epsilon=0$, the cross-entropy is independent of $n$ and, as argued above (\cref{thm:layernorm}), can be made as low as desired.

\section{Learnability}
\label{sec:learnability}

In this section, we turn to the question of learnability, which will lead to a third way of overcoming the limitation suggested by Hahn's lemma.

\subsection{Experiments: standard transformers}

We tried training transformers on both \textsf{PARITY} and \text{FIRST}. Each transformer had the same number of layers and heads and the same fixed positional encodings as the corresponding exact solution. We used $d_\text{model} = 16$ for word encodings, self-attention, and FFNN outputs, and $d_\text{FFNN} = 64$ for FFNN hidden layers. We used layer normalization ($\epsilon=10^{-5}$) after residual connections. We used PyTorch's default initialization and trained using Adam \citep{kingma+ba:2015} with learning rate $3 \times 10^{-4}$ \citep{karpathy:2016}. We did not use dropout, as it did not seem to help.

\begin{figure}
\centering
\input{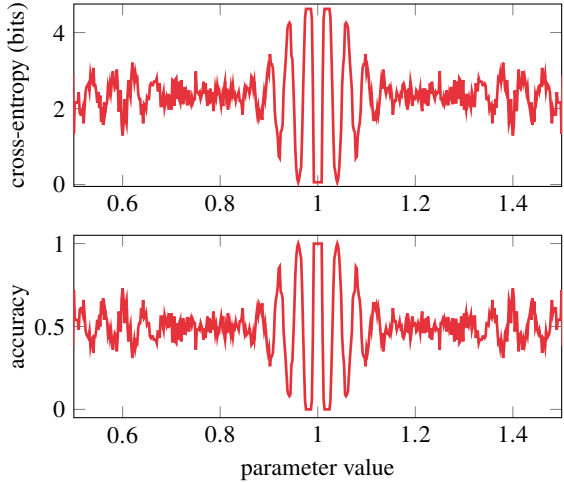}
\vspace*{-3ex}
\caption{The cross-entropy and accuracy of our solution to \textsf{PARITY} are both extremely sensitive to the parameter $\sawt[\bar]11V_{6,2}$, which is responsible for computing~$\frac{k}{n}$. The correct parameter value is 1.}
\label{fig:parity_surface}
\end{figure}

We found, like \citet{bhattamishra+:2020}, that a transformer with the above settings was unable to learn \textsf{PARITY}. We tried many other settings as well, to no avail.
To give an idea of why our constructed solution, in particular, is difficult to find, \cref{fig:parity_surface} shows the cross-entropy and accuracy of the model if we start with our solution (with layer normalization, $\epsilon=0$) and vary the parameter $\sawt[\bar]11V_{6,2}$, which is responsible for computing $\frac{k}{n}$. At~1, it has a cross-entropy of 0 and accuracy of~1, which are both optimal, but the cross-entropy oscillates so rapidly that even a small perturbation of this parameter would make it difficult to recover the solution by gradient descent.

\textsf{FIRST} is much easier to learn, but the bad news is that the learned transformers do not generalize well to longer sentences.
\Cref{fig:first_baseline} (left column) shows that when a transformer is trained from scratch on shorter strings ($n=10, 30, 100, 300$) and tested on longer strings ($n=1000$), the accuracy is not perfect. Indeed, for training $n=10$, the accuracy is hardly better than random guessing.

\subsection{Flawed transformer for \textsf{FIRST}}
\label{sec:first_bad}

In our solution above (\cref{sec:first_exact}), the second self-attention layer attended mostly to the first position, but not totally. It relied on the fact that in the second self-attention layer, the values of the non-first positions ($\Vmat^{2,1}_{i,4}$ and $\Vmat^{2,1}_{i,5}$ for $i\neq1$) are exactly zero and therefore do not contribute to the output.

In practice, because word embeddings are randomly initialized in all dimensions, and are added to every layer via residual connections, it's unlikely for any activation to be exactly zero. This explains why our exact solution cannot be learned.

But, as a further thought experiment about what the model might be learning instead, consider the following transformer, which uses only a single layer ($L=1$) and does not zero out the values of the non-first positions. As we will see, it performs worse than the transformer of \cref{sec:first_exact} for long strings.
\begin{align*}
\sawt11Q &= 
\begin{bmatrix}
0 & 0 & c\sqrt{d} & 0
\end{bmatrix}
\\
\sawt11K &=
\begin{bmatrix}
0 & 0 & 0 & 1
\end{bmatrix}
\\
\sawt11V &=
\begin{bmatrix}
\mathbf{0}^{4\times4} \\
\begin{matrix}
-\tfrac12 & \tfrac12 & -\tfrac12 & 0
\end{matrix}
\end{bmatrix}.
\end{align*}
The FFNN doesn't do anything ($\ffwt11 = \ffbias11 = \ffwt12 = \ffbias12 = \mathbf{0}$), and the final output layer just selects component 5.
So if $k$ is the total number of $\texttt{1}$s, the final logit at \texttt{CLS} (position $0$) would be
\begin{align*}
s &= \begin{aligned}[t]
&\frac{\exp c - 1}{\exp c+n-1}\left(\indicator{w_1 = \texttt{1}} - \frac12\right) \\ &{} + \frac{1}{\exp c+n-1} \left(k - \frac{n}2\right).
\end{aligned}
\end{align*}
If $c > \log (n-1)$, then this is positive iff $w_1 = \texttt{1}$. But if $c \leq \log (n-1)$, the new second term can be big enough to make the model output an incorrect answer. 
This suggests that if we train a transformer on strings with length up to $N$, then the learned parameters will be large enough to classify strings of length up to $N$ correctly, but may misclassify strings longer than $N$.

\begin{figure*}[t]
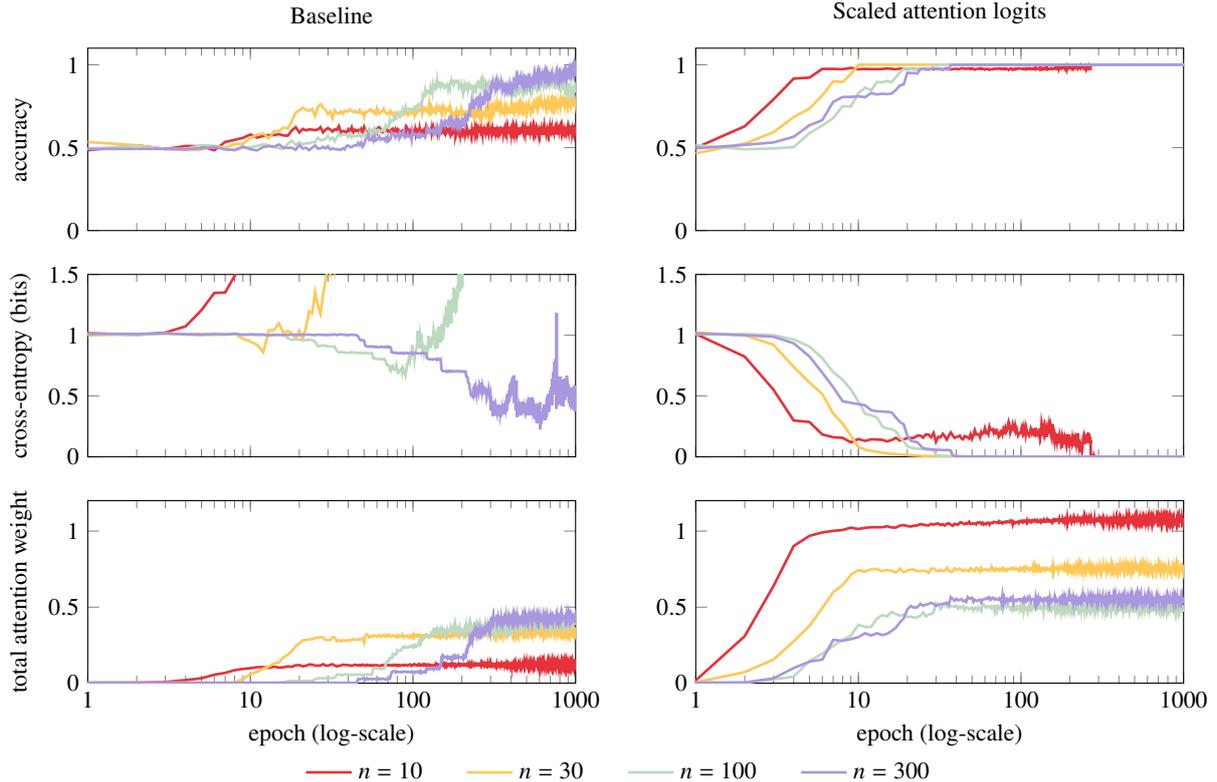
 \small \centering
\pgfplotsset{every axis/.style={xmin=1,xmax=1000,xmode=log,ymin=0,legend columns=-1,legend style={draw=none,/tikz/every even column/.append style={column sep=0.5cm}}}}
\newcommand{\axsty}[1]{\pgfplotsset{every axis/.append style={#1}}}
\begin{tikzpicture}
{\axsty{xticklabels={,,}}
{\axsty{at={(0cm,6cm)},title={Baseline}} \input{first_baseline_acc}}
{\axsty{at={(0cm,3cm)}} \input{first_baseline_ce}}}
{\axsty{at={(0cm,0cm)}} \input{first_baseline_att}}

{\axsty{xticklabels={,,}}
{\axsty{at={(8cm,6cm)},title={Scaled attention logits}} \input{first_scaled_acc}}
{\axsty{at={(8cm,3cm)}} \input{first_scaled_ce}}}
{\axsty{at={(8cm,0cm)}} \input{first_scaled_att}}
\node[anchor=north] at (7cm,-0.8cm) {\pgfplotslegendfromname{legend}};
\end{tikzpicture}
\caption{Training a transformer on \textsf{FIRST}. Each epoch has 100 training strings of varying length (see legend) and 100 test strings of length 1000. All curves are averaged over 20 runs. Left: Standard transformer with layer normalization ($\epsilon=10^{-5}$). Right: Same, with attention logits scaled by $\log n$.}
\label{fig:first_baseline}
\end{figure*}

This explanation is corroborated by the bottom-left graph in \cref{fig:first_baseline}, which shows the attention weight on the first position of the test string (summed over layers, averaged over strings) as a function of training epoch (starting from random initial parameters). The training strings have varying length ($n$) and the test strings have fixed length (1000). We might hope that the attention weight would converge to the same value independent of~$n$. But the lower $n$ is, the more the attention weight is diluted, making it easier for the value in position~1 to be outweighed by values in other positions.

\subsection{Log-length scaled attention}
\label{sec:scaled}

Fortunately, this problem is easy to fix by scaling the logits of each attention layer by $\log n$, that is, redefining attention as
\begin{equation}
\att(\mathbf{q}, \mathbf{K}, \mathbf{V}) = \mathbf{V}^\top \operatorname{softmax} \frac{\log n}{\sqrt{d}}\mathbf{K}\mathbf{q} . \label{eq:scaled_att}
\end{equation}
Then taking the model in \cref{sec:first_bad} with $c=1$ gives
\begin{align*}
s &= \frac{n-1}{2n-1}\left(\indicator{w_1 = \texttt{1}} - \frac12\right) + \frac{1}{2n-1} \left(k - \frac{n}2\right)
\end{align*}
which is positive iff $w_1 = \texttt{1}$. Moreover, scaling is another way to make the cross-entropy low: 
\begin{proposition}
For any $\eta > 0$ there is a transformer with attention defined as in \cref{eq:scaled_att}, and with or without layer normalization, that recognizes \textsf{FIRST} with cross-entropy at most $\eta$.
\end{proposition}

\begin{proof}
Without layer normalization, we can take the model in \cref{sec:first_exact}, set $c=1$ and log-scale the attention weights, which changes $s$ from \cref{eq:first_exact_logit} to
\begin{gather*}
s = \frac{n}{2n-1}\left(\indicator{w_1=\texttt{1}}-\frac12\right) \\
\frac14 < |s| \leq \frac12.
\end{gather*}

With layer normalization (and $\epsilon>0$), we can apply the modification of \cref{sec:layernorm} to nullify the centering effect of layer normalization. Then since the variance of $\avec^{2,0}$ is $\tfrac16(1+s^2)$, the layer-normalized final logit is
\begin{align*}
\bar s &= s \left(\frac16(1+s^2) +\epsilon\right)^{-\frac12}
\intertext{and since $|s|>\tfrac14$,}
|\bar s| &> \frac14 \left(\frac{5}{24} + \epsilon\right)^{-\frac12}.
\end{align*}

In either case, since the final logit has a lower bound not dependent on $n$, the output layer weights can be scaled as in the proof of \cref{thm:layernorm} to make the cross-entropy at most $\eta$.
\end{proof}

\begin{table} %\small
\begin{center}
\begin{tabular}{lll}
\toprule
%system & \multicolumn{2}{c}{setting} \\
& train all & train short \\
& test all & test long \\
\midrule
train tokens & 3M+3M & 1M+1M \\
test tokens & 32k+34k & 24k+25k \\
\midrule
baseline & 32.6 & 11.4 \\
scaled & 32.5 & 12.4 \\
\bottomrule
\end{tabular}
\end{center}
\caption{When training and testing on data with the same length distribution, scaling attention logits has no significant effect on BLEU, but when training on short sentences ($\leq 20$ tokens) and testing on long sentences ($> 20$ tokens), scaling helps significantly ($p < 0.01$).}
\label{tab:mt}
\end{table}

\subsection{Experiments: scaled attention}

\Cref{fig:first_baseline} (right column) shows the training of transformers with scaling of attention logits by $\log n$. For all training lengths $n$, the model is able to learn with perfect test cross-entropy and accuracy.

We see a similar effect on low-resource English-to-Vietnamese machine translation (\cref{tab:mt}), using Witwicky, an open-source implementation of transformers.\footnote{\url{https://github.com/tnq177/witwicky}} We use all default settings; in particular, residual connections come after layer normalization ($\epsilon=10^{-5}$).
We measure translation accuracy using BLEU \citep{papineni+:2002} and use bootstrap resampling with 1000 samples for significance testing.

When train and test length distributions are the same, scaling attention logits has no significant effect. But if we train only on sentences with  median length or shorter ($\leq 20$ tokens) and test only on sentences longer than median length ($>20$ tokens), scaling attention logits by $\log n$ improves BLEU by $+1$, which is statistically significant ($p < 0.01$).

\section{Related Work}

Using very different assumptions on the form of transformers and inputs, a number of recent theoretical studies of transformers show that they can solve much more difficult problems than the ones studied here. 
Transformer encoders can be shown to be universal approximators by fixing the string length and using a number of layers exponential in the length \citep{universal-approx-yun-etal-2020}. Transformer encoder--decoders, where the decoder can run for an unbounded number of steps, have been shown to be Turing-complete \citep{bhattamishra-etal-2020-computational,attention-turing-complete-perez-2021}.

RASP \citep{weiss+:2021} is a simple programming language whose programs can be compiled into transformers. While \textsf{PARITY} can easily be written in RASP, this does not imply in itself the existence of transformers that can recognize \textsf{PARITY}, for two reasons. First, RASP's aggregate operation (which corresponds to attention) always attends uniformly to a subset of positions, unlike softmax attention. Second, RASP's elementwise operations are embedded directly in the output transformer; they are not translated into FFNNs.

\Citet{bhattamishra+:2020} carry out theoretical and experimental studies of transformers for various formal languages. The theoretical results are for a different variant of transformers than ours (transformer encoders with self-attention masked so that each position attends only to previous positions), and focus on such transformers' ability to maintain counters that are constrained to be nonnegative. Their experimental results suggest that transformers have difficulty learning some regular languages, including \textsf{PARITY}.

\section{Conclusion}

We've seen that the questions of (a)~whether a neural network can recognize a language, (b)~whether it can achieve low cross-entropy on a language, and (c)~whether it can learn to recognize a language are three separate questions, because we have given examples of (a) without (b) and (b) without (c). 

Namely, our explicit construction for \textsf{PARITY} shows that a neural network can recognize a language with perfect accuracy (a) but poor cross-entropy (b).
Adding layer normalization ($\epsilon=0$) enables it to achieve low cross-entropy (b), but still does not learn well (c).
We observe that because the answer to (b) can hinge on small details of the model, (b) is not probably not very useful as a way of measuring expressivity. 

However, we did find that the limited influence of a single input symbol, implied by Hahn's lemma, has a serious practical implication for learnability (c). Namely, transformers can fail to generalize from shorter training strings to longer testing strings. Our proposed fix, scaling attention logits by $\log n$, is easy to implement and very effective on a real machine translation task.

\section*{Acknowledgements}

We would like to thank Toan Nguyen for assistance with his machine translation code, and Gail Weiss for catching some mistakes.

This paper is based upon work supported in part by the Office of the Director of National Intelligence (ODNI), Intelligence Advanced Research Projects Activity (IARPA), via contract \#FA8650-17-C-9116. The views and conclusions contained herein are those of the authors and should not be interpreted as necessarily representing the official policies, either expressed or implied, of ODNI, IARPA, or the U.S. Government. The U.S. Government is authorized to reproduce and distribute reprints for governmental purposes notwithstanding any copyright annotation therein.

\bibliography{references}

\clearpage\appendix

\section{Correctness of \textsf{PARITY} Construction}
\label{app:parity_details}

In \cref{sec:parity_exact}, we constructed a transformer that recognizes \textsf{PARITY}; here we fill in details of calculating $s = \avec^{2,0}_9$.
If $n$ is even, the first head computes
\begin{align*}
\qvec^{2,1,0} &= c \sqrt{d} \\
\Kmat^{2,1,0}_{i,1} &= -\cos i\pi = (-1)^{i+1} \\
\alpha^{2,1,0}_i &= \frac{\exp (-1)^{i+1} c}{\frac{n}2 (\exp c + \exp -c)} \\
\Vmat^{2,1,0}_{i,9} &= \frac{\indicator{i=k}}{n}. \\
\intertext{Similarly, the second head computes}
\qvec^{2,2,0} &= c \sqrt{d} \\
\Kmat^{2,2,0}_{i,1} &= \cos i\pi = (-1)^i \\
\alpha^{2,2,0}_i &= \frac{\exp (-1)^i c}{\frac{n}2 (\exp c + \exp -c)} \\
\Vmat^{2,2,0}_{i,9} &= -\frac{\indicator{i=k}}{n}.
\end{align*}
Then
\begin{align*}
s = \avec^{2,0}_9 &= \frac1n \alpha^{2,1,0}_k - \frac1n \alpha^{2,2,0}_k \\
&= \frac{\exp (-1)^{k+1} c - \exp (-1)^{k} c}{\frac{n^2}2(\exp c + \exp -c)} \\
&= (-1)^{k+1} \frac{\exp c - \exp -c}{\frac{n^2}2(\exp c + \exp -c)} \\
&= (-1)^{k+1} \frac{2 \tanh c}{n^2}
\end{align*}
is negative if $k$ is even and positive if $k$ is odd.

If $n$ is odd, calculating $s$ is more complicated because there are unequal numbers of more- and less-attended positions. The attention weights are
\begin{align*}
\alpha^{2,1,0}_i &= \frac{\exp (-1)^{i+1} c}{\underbrace{\textstyle \frac{n-1}2 \exp c + \frac{n+1}2 \exp -c}_{Z_1}} \\
\alpha^{2,2,0}_i &= \frac{\exp (-1)^i c}{\underbrace{\textstyle \frac{n+1}2 \exp c + \frac{n-1}2 \exp -c}_{Z_2}} \\
s %&= \frac1n \alpha^{2,1,0}_k - \frac1n \alpha^{2,2,0}_k \\
&= \frac{ (\exp (-1)^{k+1} c) Z_2 - (\exp (-1)^k c) Z_1}{nZ_1Z_2}.
\end{align*}
If $k$ is even,
\begin{align*}
s &= \frac{\frac{n-1}2 \exp -2c - \frac{n-1}2 \exp 2c}{nZ_1Z_2} \\ &= -\frac{(n-1) \sinh 2c}{nZ_1Z_2} < 0
\intertext{whereas if $k$ is odd,}
s &= \frac{\frac{n+1}2 \exp 2c - \frac{n+1}2 \exp -2c}{nZ_1Z_2} \\ &= \frac{(n+1) \cosh 2c}{nZ_1Z_2}> 0.
\end{align*}

\section{Scale-Invariance of \textsf{PARITY} and \textsf{FIRST} Constructions}
\label{app:scale}

In \cref{sec:recentering}, we claimed that the scaling effect of layer normalization has no effect on the decisions of our constructions for \textsf{PARITY} and \text{FIRST}.
This is related to the property of approximate homogeneity studied by \citet{merrill+:2021}.

In general, we rely on the fact that the FFNNs we use all have no bias terms ($\ffbias\ell1$ and $\ffbias\ell2$), so the FFNNs are 1-homogenous (scaling the input scales the output by the same amount). For the self-attentions, our $\sawt{\ell}hQ$ all have a constant factor $c$ built into them, so any scaling of the input can be absorted into this constant.

For \textsf{PARITY}, suppose that layer normalization scales $\cvec^\ell$ by $C_\ell$ and $\avec^\ell$ by $A_\ell$.
\begin{align*}
\bar\cvec^{1,i} &= C_1 \begin{bmatrix}\cvec^{1,i} \\ -\cvec^{1,i} \end{bmatrix} \\
\intertext{Because the first FFNN has no bias term,}
\bar\avec^{1,i} &= A_1C_1 \begin{bmatrix}\avec^{1,i} \\ -\avec^{1,i} \end{bmatrix}
\end{align*}
In the second self-attention layer, the attention logits and the values are scaled by $A_1C_1$.
We're only interested in what happens to $s = \cvec^{2,0}_9$. If $n$ is even, $s$ becomes:
\begin{equation*}
\bar{s} = (-1)^{k+1} \frac{2 C_2 A_1 C_1 \tanh A_1 C_1 c}{n^2}.
\end{equation*}
Since the second FFNN is the identity function, its layer normalization has no effect ($A_2=1$).
So the final logit is $\bar{s}$, which is still negative if $k$ is even and positive if $k$ is odd. Similarly if $n$ is odd.

For \textsf{FIRST}, again suppose that layer normalization scales $\cvec^\ell$ by $C_\ell$ and $\avec^\ell$ by $A_\ell$. As before,
\begin{align*}
\bar\avec^{1,i} &= A_1C_1 \begin{bmatrix}\avec^{1,i} \\ -\avec^{1,i} \end{bmatrix}
\end{align*}
In the second self-attention layer, the attention logits and the values are scaled by $A_1C_1$.
We're only interested in what happens to $s = \cvec^{2,0}_6$:
\begin{equation*}
\bar{s} = \frac{\exp A_1C_1 c}{\exp A_1C_1 c + n - 1} C_2A_1C_1 \left(\indicator{w_1 = \texttt{1}} - \tfrac12\right)
\end{equation*}
Since the second FFNN is the identity function, $A_2=1$. So the final logit is $\bar{s}$, which is still positive if $w_1 = \texttt{1}$ and negative otherwise.

\end{document}